\documentclass[runningheads]{llncs}
\usepackage{makeidx}
\usepackage{graphicx}
\usepackage{amsmath,amssymb}
\usepackage{color}

\usepackage{booktabs}
\usepackage{tabularx}
\usepackage{wrapfig}
\usepackage[format=plain,labelformat=simple,labelsep=period,font=small,skip=4pt,compatibility=false]{caption}
\usepackage[font=scriptsize,skip=2pt]{subcaption}
\usepackage[pagebackref=true,breaklinks=true,letterpaper=true,colorlinks,bookmarks=false]{hyperref}

\usepackage{multibib}
\newcites{supp}{References}

\usepackage[capitalize]{cleveref}
\crefname{section}{Sec.}{Secs.}

\usepackage{balance}
\usepackage{fancyhdr}

\clearpage{}
\makeatletter
\DeclareRobustCommand\onedot{\futurelet\@let@token\@onedot}
\def\@onedot{\ifx\@let@token.\else.\null\fi\xspace}

\def\eg{\emph{e.g}\onedot}
\def\ie{\emph{i.e}\onedot}
\def\cf{\emph{c.f}\onedot}
\def\wrt{w.r.t\onedot} 
\def\etal{\emph{et al}\onedot}
\makeatother

\usepackage{xspace}
\newcommand{\Eq}{Eq.\@\xspace}

\newcommand{\R}{\ensuremath{\mathbb{R}}}
\newcommand{\Z}{\ensuremath{\mathbb{Z}}}

\newcommand{\mv}[1]{\ensuremath{\mathbf{#1}}}
\newcommand{\mm}[1]{\ensuremath{\mathbf{#1}}}

\newcommand{\f}{\mathbf{f}}
\newcommand{\p}{\mathbf{p}}
\renewcommand{\o}{\mathbf{o}}
\newcommand{\vv}{\mathbf{v}}
\newcommand{\ii}{\mathbf{i}}
\newcommand{\btheta}{\boldsymbol{\theta}}

\newcommand{\myparagraph}[1]{\smallskip\noindent\textbf{#1}\hspace{0.5em}}

\usepackage{pifont}
\newcommand{\cmark}{\ding{51}}
\newcommand{\xmark}{\ding{55}}\clearpage{}

\begin{document}
\pagestyle{headings}
\mainmatter

\title{Learning Task-Specific Generalized Convolutions in the Permutohedral Lattice}
\titlerunning{Learning Task-Specific Generalized Convolutions in the Perm. Lattice}
\authorrunning{Anne S. Wannenwetsch, Martin Kiefel, Peter V. Gehler, Stefan Roth}
\author{
Anne S. Wannenwetsch\inst{1}\thanks{This project was mainly done during an internship at Amazon, Germany.}\orcidID{0000-0002-7016-3820}, \\
Martin Kiefel\inst{2}\orcidID{0000-0001-9432-5428}, \\
Peter V. Gehler\inst{2}\orcidID{0000-0002-5812-4052},
Stefan Roth\inst{1}\orcidID{0000-0001-9002-9832}
}
\institute{$^1$ TU Darmstadt, Germany \qquad $^2$ Amazon, Germany}

\maketitle
\thispagestyle{fancy}

\begin{abstract}
	
Dense prediction tasks typically employ encoder-decoder architectures, but the prevalent convolutions in the decoder are not image-adaptive and can lead to boundary artifacts.
Different generalized convolution operations have been introduced to counteract this.
We go beyond these by leveraging guidance data to redefine their inherent notion of \emph{proximity}.
Our proposed network layer builds on the \emph{permutohedral lattice}, which performs sparse convolutions in a high-dimensional space allowing for powerful non-local operations despite small filters.
Multiple features with different characteristics span this permutohedral space.
In contrast to prior work, we \emph{learn} these features in a task-specific manner by generalizing the basic permutohedral operations to learnt feature representations. 
As the resulting objective is complex, a carefully designed framework and learning procedure are introduced, yielding rich feature embeddings in practice.
We demonstrate the general applicability of our approach in different joint upsampling tasks.
When adding our network layer to state-of-the-art networks for optical flow and semantic segmentation, boundary artifacts are removed and the accuracy is improved. \end{abstract}

\section{Introduction}
Deep learning approaches are the backbone of many state-of-the-art methods across computer vision \cite{Chen:2018:EAS,Singh:2018:SEM,Sun:2018:PCO}.
Convolutional neural networks (CNNs) are particularly common as they greatly lower the number of parameters compared to fully-connected networks and thus scale to practically relevant image sizes.
While early CNNs employed large filters \cite{Krizhevsky:2012:ICD}, it is now common to use small kernels stacked into deep networks \cite{He:2017:DRL,Simonyan:2015:VDC}.
Chaining several smaller filters requires fewer parameters for the same receptive field of a single large filter, and leads to more discriminative features by virtue of having more non-linearities \cite{Simonyan:2015:VDC}.

While convolutions build a fundamental block of deep learning, they are not without drawbacks. 
First, they are not image-adaptive, \ie content boundaries in a feature map are not respected but smoothed over.
This is especially disadvantageous for dense prediction tasks, \eg semantic segmentation or optical flow, leading to accuracy loss at boundaries \cite{Harley:2017:SCN,Wu:2018:DFL}.
Moreover, convolutions have a limited and predefined receptive field, which connects spatially close regions but cannot leverage similar, but more distant image structures.
Here, a new definition of \emph{pixel proximity} is needed that goes beyond two-dimensional (2D) spatial distance.
For instance, image values themselves or abstract properties such as object classes could be used to define similarity in a more general setting.

Several methods have been proposed to counteract the named disadvantages.
Sampling-based approaches \cite{Jaderberg:2015:STN,Recasens:2018:LZS} rearrange the image content
but remain restricted to the 2D concept of proximity.
Location specific networks \cite{Jia:2016:DFN,Wu:2018:DFL} predict pixelwise filter kernels, but require many additional parameters.
Other methods \cite{Dai:2017:DCN,Tabernik:2018:SFU} determine neighboring pixels in an image-adaptive manner.
However, the convolutional structure is fixed and only the position of neighbors is adjustable.

Image-adaptive filters, \eg \cite{He:2010:GIF,Tomasi:1998:BFG}, have been used in traditional computer vision for years.
The bilateral filter \cite{Tomasi:1998:BFG} adapts a Gaussian kernel according to the spatial distance and color difference of neighboring pixels.
In \cite{Jampani:2016:LSH,Kiefel:2015:PLC}, this concept is leveraged to construct bilateral convolution layers (BCLs) based on the \emph{permutohedral lattice} \cite{Adams:2010:FHF} -- a fast approximation of the bilateral filter.
Filtering corresponds to a sparse convolution in a high-dimensional space, which is spanned by different features, \eg spatial location and color.
Jampani \etal \cite{Jampani:2016:LSH,Kiefel:2015:PLC} extend the Gaussian kernel to a general, image-adaptive convolution and learn the kernels from data.
However, the features constituting the lattice space remain fixed.
Feature parameters are not adjustable during training, which complicates integration into end-to-end learning.
More importantly, relying on predefined features without further processing omits a possible source of improvements.

To counteract this disadvantage of BCLs, we present the \emph{semantic lattice layer}.
We rely on the permutohedral lattice as a backbone and show how to generalize its operations \wrt features with learnable parameters.
The resulting computations are involved and may lead to practical challenges.
We hence propose a specific setting in which basic features -- as used in \cite{Jampani:2016:LSH,Kiefel:2015:PLC} -- are processed by a CNN.
This greatly simplifies the optimization since it allows to combine and especially refine features that are known to be beneficial for image-adaptive filtering.
We further present various measures to avoid difficulties during learning.
For instance, as the sparsity of the semantic lattice may avoid propagation of information if pixels are too distant, we restrict the output range of the embedded features.
This rather simple measure has a large effect in practice.

Our setup enables us to learn meaningful feature embeddings from data.
It allows to integrate feature parameters into training and effectively leverages guidance data to connect pixels due to their similar characteristics.
As such, the semantic lattice is able to perform non-local operations 
while keeping the filter kernels and consequently the number of learnt parameters small and manageable.

We show the benefits of the semantic lattice in different areas for image-adaptive upsampling.
For the task of color upsampling, the semantic lattice outperforms previous approaches by a large margin.
We further replace bilinear upsampling in state-of-the-art networks for optical flow and semantic segmentation.
Here, the semantic lattice leads to better aligned and crisper content boundaries and also improves the accuracy, especially at discontinuities. 
\section{Related Work}
\subsubsection{Generalized Convolutions.}
We begin by reviewing work that generalizes convolution operations.
Jaderberg \etal  \cite{Jaderberg:2015:STN} introduce Spatial Transformers (STs), which transform feature maps depending on the data itself.
Similar to warped convolutions \cite{Henriques:2017:WCE}, STs aim for invariance to certain transformations, \eg rotation or scaling.
\cite{Li:2017:DTN} applies STs to allow for irregular patches in dense prediction tasks.
In \cite{Recasens:2018:LZS}, saliency-based sampling emphasizes regions of high interest.
These methods rearrange data in 2D space.
In contrast, we can 
leverage additional feature dimensions to redefine the concept of pixel proximity.

Filter-weight networks \cite{Jia:2016:DFN} generate location-specific filters dependent on the input image.
In \cite{Kang:2017:ISI}, adaptive weights incorporate side information about the scene context.
However, adaptive filters introduce many additional parameters in comparison to location-invariant networks
and remain restricted to local transformations due to a fixed receptive field.
Wu \etal \cite{Wu:2018:DFL} apply location-specific convolutions not only to the position itself but to several sampled neighboring regions, which extends their receptive field but requires additional computations.

Dilated convolutions use a fixed spacing between considered pixels to extend the spatial resolution \cite{Chen:2015:SIS,Yu:2016:MCA}.
It is possible to learn offsets for the input locations of each filter \cite{Jeon:2017:ACL} or have them depend on the input and spatial location \cite{Dai:2017:DCN}.
When using mixtures of Gaussians as filters, size and location of the receptive fields are learnable \cite{Tabernik:2018:SFU}.
Structure-aware convolutions \cite{Chang:2018:SCN} use univariate functions as filters and are also applicable to non-Euclidean data. 
We do not learn individual neighborhoods for all filters but convolutions are instead performed consistently in a learnt feature space.
Moreover, our convolution structure is not fixed; the number of neighbors is flexible and homogenous areas can be compressed.

\myparagraph{Permutohedral Lattice.}
Adams \etal \cite{Adams:2010:FHF} propose the permutohedral lattice as a fast method for high-dimensional Gaussian filtering.
It found widespread application, 
especially for fast inference in dense Conditional Random Fields (CRFs) \cite{Kraehenbuehl:2011:EIF,Perazzi:2012:SFC,Zhang:2016:ISA}
and upsampling or densification of data \cite{Dolson:2010:URD,Russell:2014:VPM}.
In contrast to our work, these approaches use fixed Gaussian filters and predefined features.
\cite{Kraehenbuehl:2013:PLC} extends the fast inference method of \cite{Kraehenbuehl:2011:EIF} to learn parameters of dense CRFs, but the setup is restricted to Gaussian filters and customized to its application.

In \cite{Jampani:2016:LSH,Kiefel:2015:PLC}, the high-dimensional filtering in permutohedral space is generalized by learnable convolution parameters.
The proposed BCLs are beneficial in neural networks as they allow to redefine proximity of pixels \wrt different characteristics \cite{Gadde:2016:SCN,Jampani:2016:LSH,Li:2018:UVO}.
Moreover, BCLs can inherently cope with sparse data \cite{Kiefel:2015:PLC}, \eg in 3D point cloud processing \cite{Su:2018:SSL}.
Again, all methods rely on predefined features and thus restrict the flexibility of the generalized convolutions.
We will show that a general setup with learnt features leads to better results in practice.

Another line of research aims for further speed-up of the permutohedral lattice. 
For instance, \cite{Dai:2018:DTA} proposes to encode its operations in a deep neural net.

\myparagraph{Learnt Representations.}
Our embedding network aims to encode guidance data as discriminatively as possible for the task at hand. 
As such, our work is related to general embedding or metric learning;  
see \cite{Song:2016:DML,Wu:2017:SMD} for an overview.

\myparagraph{Image-Adaptive Filtering.}
We leverage additional properties for our generalized convolutions, which closely relates our approach to image-adaptive filtering methods
such as bilateral filtering \cite{Tomasi:1998:BFG}, non-local means \cite{Buades:2005:NLA}, and guided image filtering \cite{He:2010:GIF}.
All of these filters have been included into deep networks, 
\eg for semantic segmentation \cite{Gadde:2016:SCN,Harley:2017:SCN}, image processing \cite{Wu:2018:FET}, or video classification \cite{Wang:2018:NNN}.

Only few approaches aim to learn guidance features for the filtering step in a general context.
Harley \etal \cite{Harley:2017:SCN} propose segmentation-aware convolutions, 
which leverage image-adaptive masks from an embedding network.
Object class labels are required for pre-training and large filter kernels increase the risk of overfitting.
Deep joint image filtering (DJIF) \cite{Li:2019:JIF} uses two individual networks to preprocess guidance and data features and subsequently merges the two branches for joint filtering.
However, explicit knowledge about the relation between guidance and target data is not leveraged.
Gharbi \etal~\cite{Gharbi:2017:DBL} reproduce image enhancement operators with locally-affine models and upsample the low-resolution outputs guided by a learnt feature channel.
Here, the offline learning of the models puts strong restrictions on the approximated operators.  
Deep guided filters \cite{Wu:2018:FET} allow to learn a guidance image but restrict its dimensionality to a one-dimensional signal per output channel.

In contrast to previous work, the semantic lattice is applicable to a large variety of tasks and puts no restrictions on the guidance data.
Moreover, the rich feature representations allow us to keep the applied filter kernels small. 
\section{The Semantic Lattice}
To allow for the non-local combination of data, we build on the permutohedral lattice \cite{Adams:2010:FHF} to redefine the notion of \emph{proximity} between the pixels of an image.
The permutohedral lattice assumes that each input point is characterized by two properties -- \emph{features} and \emph{data}.
The feature value $\f \in \R^d$ indicates the location of the respective pixel in the $d$-dimensional permutohedral space, while the data value $\vv \in \R^c$ describes the information stored at this location.
In a first step, the data is projected into the lattice grid using the features to determine its position.
Convolutions can then be performed in permutohedral space, considering a neighborhood defined by the feature dimensions. 
For instance, color values can be considered to connect visually similar areas and respect object boundaries \cite{Jampani:2016:LSH,Kiefel:2015:PLC}.
This is in contrast to regular convolutions where the spatial location is used as the only feature to determine neighboring pixels. 
Finally, the convolved output is extracted at certain feature positions, which can but do not have to coincide with the input locations depending on the task at hand.

In the following, we introduce the permutohedral lattice and its properties more formally. 
We then extend the work in \cite{Jampani:2016:LSH,Kiefel:2015:PLC} to eliminate the usage of fixed, hand-crafted features.
In particular, we show how to learn an appropriate feature space based on spatial positions as well as guidance data.
As this approach allows to leverage semantically meaningful properties that go beyond the concept of predefined features, we refer to our proposed setup as the \emph{semantic lattice}.

\subsection{Permutohedral Lattice}
Following \cite{Adams:2010:FHF}, the permutohedral lattice is defined as the projection of the regular grid $\Z^{d+1}$ onto the hyperplane $\mathcal{H}: \mv{h} \cdot \mv{1} = 0 \subseteq \R^{d+1}$.
The projected grid points thus represent the corners of permutohedral simplices, which split the hyperplane $\mathcal{H}$ into uniform cells.  
We refer to \cite{Adams:2010:FHF} for further details of the lattice structure.

\begin{figure}[t]
\centering
	\begin{subfigure}[b]{0.25\linewidth}
		\includegraphics[width=\linewidth]{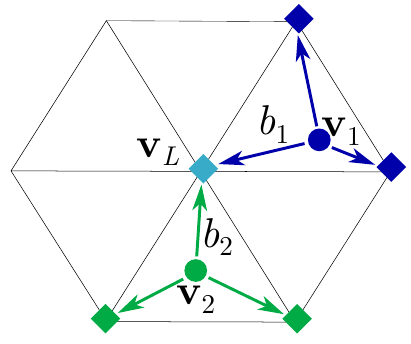}
		\caption{Splat.}
		\label{fig:permutohedral_lattice_splat}
	\end{subfigure}
	\hfill
	\begin{subfigure}[b]{0.25\linewidth}
		\includegraphics[width=\linewidth]{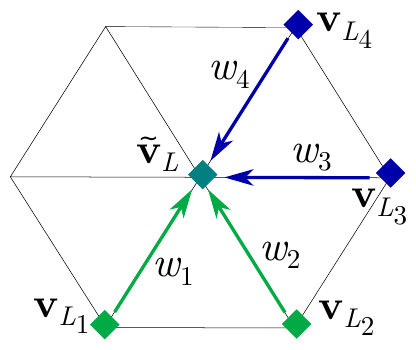}
		\caption{Convolve.}
		\label{fig:permutohedral_lattice_convolve}
	\end{subfigure}
	\hfill
	\begin{subfigure}[b]{0.25\linewidth}
    	\centering
   		\includegraphics[width=\linewidth]{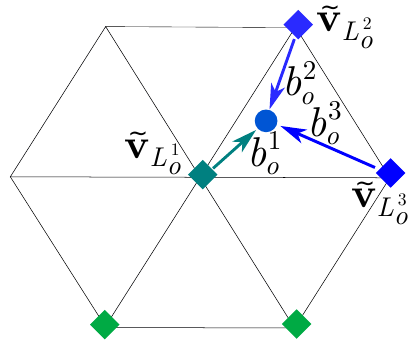}
   		\caption{Slice.}
   		\label{fig:permutohedral_lattice_slice}
 	\end{subfigure}
\caption{Basic operations in the permutohedral lattice.}
\label{fig:permutohedral_lattice}
\vspace{-0.5em}
\end{figure}

The operation to read data into the lattice grid is denoted as \emph{splatting}, see \cref{fig:permutohedral_lattice_splat}.
The feature vector $\f_i$ is used to place an input point $\ii = (\f_i, \vv_i)$ into the permutohedral lattice. 
Then, the data value is splat onto the enclosing lattice points according to its barycentric coordinates.
The data value of a lattice point $L$ is given as 
\begin{equation}
\vv_L = \sum_{i \in \mathcal{I}(L)} b_i \cdot \vv_i, 
\label{eq:splat_operation}
\end{equation}
where $b_i$ denote the barycentric coordinates of input points $\mathcal{I}(L)$ splatting on $L$.

The \emph{convolution} step is subsequently performed on the permutohedral grid points considering corners in a neighborhood $\mathcal{N}(L)$, \cf \cref{fig:permutohedral_lattice_convolve}. 
If a neighboring corner was not set during splatting, its value is assumed to be zero.
Using a kernel ${\mm{W} = (w_1, \ldots, w_N)}$, the convolution results in the updated lattice data
\begin{equation}
\tilde{\vv}_L = \sum_{L_n \in \mathcal{N}(L)} w_n \cdot \vv_{L_n}.
\label{eq:convolve_operation}
\end{equation}

\cref{fig:permutohedral_lattice_slice} illustrates the final \emph{slicing} operation, which interpolates the data from lattice points to an output pixel $\o$.
The value at pixel position $\f_o$ is obtained as
\begin{equation}
\tilde{\vv}_o = \sum_{k=1}^{d+1} b_o^k \cdot \tilde{\vv}_{L_o^k},
\end{equation}
with enclosing simplex corners $L_o^k$ and barycentric coordinates $b_o^k$, $1 \leq k \leq d+1$.

In \cite{Jampani:2016:LSH,Kiefel:2015:PLC}, the permuthedral lattice is integrated into deep learning 
by providing partial derivatives of the permutohedral operations \wrt the input data $\vv$ and the kernel $\mm{W}$. 
As such, the original Gaussian kernel \cite{Adams:2010:FHF} is transformed into a general convolution with a flexible filter $\mm{W}$ learnt from data.
As the filter operation is performed in lattice space, the convolution respects the notion of proximity introduced by the features $\f$ that span the permutohedral lattice.

\subsection{Generalized Features in the Semantic Lattice}
\label{sec:generalized_features}
To define the permutohedral space, \cite{Jampani:2016:LSH,Kiefel:2015:PLC} resort to predefined features, which are usually taken as $\f = (x, y, r, g, b)$.
Here, $x$ and $y$ describe the spatial x- and y-coordinates of a pixel, which are concatenated with corresponding RGB values.
                  
Our semantic lattice instead aims to learn feature embeddings from data to leverage the full capacity of the lattice.
To that end, we introduce generalized input features $\f (\ii ; \btheta_I)$ that depend on each pixel $\ii$ as well as a global set of parameters $\btheta_I$.
The splatting operation in \cref{eq:splat_operation} then generalizes to 
\begin{equation}
\vv_L(\btheta_I) = \sum_{i \in \mathcal{I}(L; \btheta_I)} b_i (\btheta_I) \cdot \vv_i,
\label{eq:generalized_splat_operation}
\end{equation}
since the splatting points as well as the corresponding barycentric coordinates depend on the feature values and thus also on $\btheta_I$.
Due to the fixed lattice structure, the set of neighbors for the convolution remains unchanged.
However, the data value $\vv_L$ at each lattice point depends on the inputs that splatted to this exact corner 
such that we rewrite the convolution in \cref{eq:convolve_operation} as
\begin{equation}
\tilde{\vv}_L (\btheta_I) = \sum_{L_n \in \mathcal{N}(L)} w_n \cdot \vv_{L_n} (\btheta_I).
\label{eq:generalized_convolve_operation}
\end{equation}
Finally, the set of lattice points surrounding an output pixel $\o$ and its barycentric coordinates are again dependent on its features $\f(\o,\btheta_O)$, 
which are parametrized by a set $\btheta_O$.
This definition results in a generalized slicing operation given as
\begin{equation}
\tilde{\vv}_o (\btheta_I, \btheta_O) = \sum_{k=1}^{d+1} b_o^k (\btheta_O) \cdot \tilde{\vv}_{L_o^k (\btheta_O)} (\btheta_I).
\label{eq:generalized_slice_operation}
\end{equation}

As operations in the lattice require specific computations, common automatic differentiation packages cannot be easily applied.
Instead, we rely on customized functions for the above generalized operations as well as their parameter gradients.
The derivatives then allow us to apply gradient based optimizers to learn task-specific feature representations $\f(\ii; \btheta_I)$ and $\f(\o; \btheta_O)$ from data.

However, the nested occurance of the parameter sets $\btheta_I$ and $\btheta_O$ already suggests that learning these generalized features may not be straightforward.
Reconsidering the generalized operations in \cref{eq:generalized_splat_operation,eq:generalized_convolve_operation,eq:generalized_slice_operation}, 
we observe that information between input and output pixels only propagates via a set of lattice corners defined by the neighborhood size of the convolution step.
It is thus essential that input and output feature positions are sufficiently close in lattice space when starting the learning process.
Otherwise, the loss gradient does not affect the input feature parameters $\btheta_I$ and no learning occurs.

\begin{figure*}[tb]
\centering
\includegraphics[width=\linewidth]{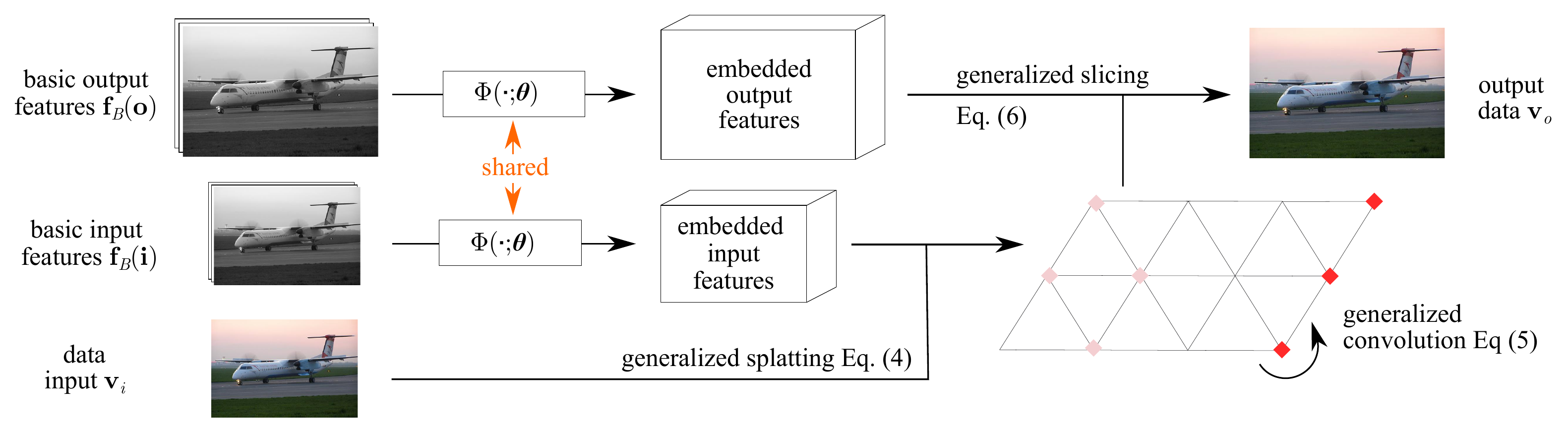}
\caption{Visualization of generalized feature learning in the semantic lattice; illustrated for the task of guided color upsampling.}
\label{fig:feature_learning}
\vspace{-0.5em}
\end{figure*}

To avoid this situation, we propose a specific framework as illustrated in \cref{fig:feature_learning} for the sample task of color upsampling.
For given input and output points $\p$, we first generate several \emph{basic features} $\f_B(\p) \in \R^{d'}$, 
\ie hand-crafted features that we assume to be helpful for the task of interest.
In the example case, the spatial location of each pixel and the corresponding grayscale image are chosen as basic features. 
Here, the additional grayscale information needs to be available for the input as well as output pixels.
We denote it more generally as \emph{guidance data} in the following.
Then, a parametric function $\Phi: \R^{d'} \rightarrow \R^d$ is defined, which takes the basic features $\f_B$ as an input and returns a learnt feature embedding $\Phi(\f_B(\p); \btheta)$ in $\R^d$.
The parameter set $\btheta$ is shared across input and output points, \ie $\btheta = \btheta_I = \btheta_O$, to ensure the necessary consistency of the feature embedding.
We propose to use a multi-layer CNN for the embedding function $\Phi$ and refer to this network in the following as \emph{feature} or \emph{embedding network}.
The parameter set $\btheta$ thus corresponds to the network weights.
Embedded features 
as well as data inputs $\vv_i$ are then used for the generalized operations in
\cref{eq:generalized_splat_operation,eq:generalized_convolve_operation,eq:generalized_slice_operation}.

As there are no restriction on the basic features, the semantic lattice is able to learn different kinds of non-local operations.
Due to the characteristics of the permutohedral lattice, input and output positions are rearranged according to the learnt features
and spatially far pixels are connected if they share the same characteristics. 
With this redefinition of proximity, the number of weight parameters remains limited while the semantic lattice yet operates globally. 
\section{Training the Semantic Lattice}

The training and setup of the semantic lattice requires careful consideration.
We detail this in the following and provide an experimental analysis in \cref{sec:ablation_study}.

\subsection{Training Procedure}
\label{sec:training_procedure}
\subsubsection{Feature Scaling.}
As the size of the permutohedral lattice cells is fixed, a scaling factor applied to the individual features determines the importance of the different dimensions as well as the number of pixels that fall into one lattice cell.
These scaling factors thus constitute important hyperparameters.
Following \cite{Jampani:2016:LSH,Kiefel:2015:PLC}, we determine factors for our basic features via grid search. 
While with the semantic lattice it is possible to refine these factors in end-to-end training through backpropagation, we found little benefit in our experiments.
Hence, we use the scaled features from the grid search as inputs to our embedding network.

\myparagraph{Data Centering.}
If predefined features are used to map to the lattice space, the output is largely invariant to a global translation of the features.
In contrast, we found feature network training to be more stable with zero mean input.
We thus subtract the dataset mean from the basic features before feature scaling.

\myparagraph{Explicit Spatial Features.}
As the embedding network combines basic features with various scale factors, we find that a random initialization may lead to a poor initial accuracy.
While the feature network is able to recover a reasonable embedding starting from a random initialization, we observe long training times in practice as well as occasional convergence to poor local minima (inferior to the scaled basic features themselves).
We find that this is mainly caused by the absence of reliable spatial coordinate features in the initial embedding network.
Hence, we do not input the spatial coordinates into the embedding network, and instead explicitly concatenate the scaled spatial features and the learnt feature embedding to jointly define the lattice space for the subsequent convolutions.

\myparagraph{Normalization Weights.}
The number of points per lattice cell can vary considerably, resulting in differing ranges of absolute data values at corner points (\Eq~\ref{eq:generalized_splat_operation}).
Moreover, the flexible structure of the lattice results in a variable number of non-zero neighbors for the convolution in \cref{eq:generalized_convolve_operation}. 
For this reason, computations in permutohedral space require a normalization step on the slice result in \cref{eq:generalized_slice_operation}.
We divide by a normalization value, which is obtained by performing all lattice operations with a placeholder input with the same features as the regular input and $\vv_i=1$.
This implies that an all-one input remains unchanged by the lattice operations.
Note that this normalization becomes invalid as convolution weights turn negative.
\cite{Jampani:2016:LSH,Kiefel:2015:PLC} resolve this by introducing a separate set of convolution weights for the normalization. 
They rely on a fixed Gaussian filter, which reduces the flexibility.
In contrast, we explicitly learn separate convolution weights for the normalization step and constrain them to be non-negative.

\myparagraph{Learning Rates.}
If the feature network and permutohedral kernels are learnt simultaneously, 
individual learning rates are applied to both parameter sets.

\subsection{Architecture}
\label{sec:training_framework}
We use a CNN with $3\times3$ filters and leakyReLU activations as our feature embedding network.
The non-linearities are omitted after the last convolution to allow for positive and negative features.\footnote{Details of the network architecture are provided in the supplemental material.}
We experimented with ResNet-like feature networks \cite{He:2017:DRL}, but observed little benefit.
In contrast, we found it essential to add a batch normalization layer \cite{Ioffe:2015:BNA} at the end of the embedding network, \cf \cref{sec:ablation_study}.
This can be understood as follows:
Even in our carefully designed semantic lattice, it is possible that no data is splat to the lattice cells surrounding a specific output location.
In such a case, the slice operation returns zero and the corresponding gradient with respect to the output location turns zero as well.
Without further gradient signals from such pixels, learning keeps pushing more pixels into this disadvantageous state and the accuracy starts to degrade.
Consequently, it is necessary to restrict the output range of the feature network, which batch normalization admits.
While other normalization methods are possible, \eg a simple min-max normalization, they show no clear benefit over batch normalization, which is commonly available in deep learning libraries.

In permutohedral space, we use a single kernel per input channel with a neighborhood of size one, \cf supplemental material.
For upsampling tasks as in \cref{sec:guided_upsampling}, 
transitions of the sparse inputs between lattice cells may cause sudden changes of training loss.
For this reason, we apply a nearest neighbor upsampling to the low-resolution guidance and input data before feeding them into the feature network and lattice, respectively.
The spatial features are adapted to the upsampled input, which spreads the data more evenly over the lattice and leads to more reliable gradients \wrt the features. 
\section{Experiments}

\subsection{Color Upsampling}
\label{sec:guided_upsampling}
Guided upsampling is a common application of image-adaptive filters \cite{Barron:2016:FBS,Kopf:2007:JBU,Li:2019:JIF,Wu:2018:FET}.
Here, guidance data is available at a higher resolution than the data of interest.
This is particularly interesting if sensor data is available at different resolutions. 

We evaluate our approach on the the task of joint color upsampling 
in which a grayscale image guides the upsamling of a low-resolution color image.
Following \cite{Jampani:2016:LSH}, we use images of the Pascal VOC Segmentation splits \cite{Everingham:2015:PVO} for training, validation, and test from which we removed grayscale images for fair comparison.
Bilinear interpolation is used to downsample color and grayscale images by $4\times$. 

The semantic lattice is applied to learn the offset between grayscale images and the RGB data.
We use spatial coordinates and grayscale values as basic features.
The semantic lattice is trained for 100 epochs on random crops of size $200 \times 272$ with
learning rates of $0.001$ and $0.01$ for the feature network and permutohedral kernels, respectively.
For comparison, we also train the deep guided filter (DGF) \cite{Wu:2018:FET} in the same setting for $150$ epochs using their procedure for image processing tasks.
Again, the DGF predicts the offset between RGB and grayscale images as this slightly improves the results.
We also compare with Deep Joint Image Filtering (DJIF) \cite{Li:2019:JIF} by applying their residual network trained for the task of depth upsampling to predict color offsets.

\cref{tab:results_colorization_upsampling} summarizes color upsampling results
on Pascal VOC Segmentation test.
\begin{table}[tb]
  \centering
  \small
  \caption{PSNR for \emph{color upsampling} on the Pascal VOC 2012 Segmentation test set.}
  \label{tab:results_colorization_upsampling}
  \smallskip
  \begin{minipage}[c]{0.58\linewidth}
  \begin{tabularx}{\textwidth}{@{}Xc@{}}
  \toprule
   & PSNR  [dB]  \\ 
  \midrule
  Semantic lattice (\textit{scaled basic features})	& 36.55	\\
  Semantic lattice (\textit{learnt kernels})				& 36.62	\\ 
  Semantic lattice (\textit{learnt embedding})		& 36.81	\\  
  Semantic lattice (\textit{both learnt})					& \bfseries 36.83	\\
  \bottomrule 
 \end{tabularx}
  \end{minipage}
  \hfill
  \begin{minipage}[c]{0.39\linewidth}
  \begin{tabularx}{\textwidth}{@{}Xc@{}}
  \toprule
   & PSNR  [dB]  \\ 
  \midrule
  Nearest neighbors				& 22.17 	\\
  Bicubic upsampling				& 23.45 	\\
  DGF \cite{Wu:2018:FET}	& 35.17	\\
  DJIF \cite{Li:2019:JIF}		& 	23.99	\\
  \bottomrule
\end{tabularx}
  \end{minipage}
\end{table}
When learning the permutohedral kernels (\textit{learnt kernels}), we observe only a small benefit in comparison to the usage of a Gaussian filter (\textit{scaled basic features}).
In contrast, our learnt feature embedding (\textit{learnt embedding}) leads to a significant improvement, highlighting the importance of using task-specific features.
Combining both leads to another (minor) gain.
Overall, we outperform the baselines of nearest neighbor and bicubic upsampling as well as related work \cite{Li:2019:JIF,Wu:2018:FET} by a large margin.
Please see supplemental material for visualizations. 
\subsection{Validation of Architectural Choices}
\label{sec:ablation_study}
We now compare different settings for feature learning to validate our proposed lattice setup.
\cref{tab:ablation_study_colorization} summarizes results obtained with fixed Gaussian kernels.
First, we train an embedding network without batch normalization to evaluate the importance of restricting its output range.
We observe a significant drop in PSNR with a result only slightly better than that with scaled basic features.
This is due to the fact that input and output locations do not necessarily coincide, which may lead to empty cells without gradients, \cf \cref{sec:training_framework}.
If the output range of the network is restricted, the number of such pixels can be kept small.

Next, we validate feeding our embedding network with guidance data and concatenating its output with spatial features.
We first learn the scale factor of x- and y-coordinates jointly with the embedding.
As this yields a negligible improvement over our baseline, we generally do not refine the scale factors.
However, note that bigger benefits may be obtained from scale refinement if the initial scale factors are estimated only coarsely.
In a second experiment, we apply the embedding
\begin{wraptable}{r}{0.5\textwidth} 
\vspace{-3px}
 \centering
  \small
  \caption{Validation of architectural choices for color upsampling on Pascal VOC test.}
  \label{tab:ablation_study_colorization}
  \begin{tabular*}{\linewidth}{lcc}
  \toprule
   & PSNR [dB]  \\
  \midrule
  Baseline	(\textit{learnt embedding}) 		& 36.81 \\
  W/o batch normalization layer	& 36.61 \\  
  Learnt spatial scale factor			& \bfseries 36.82 \\ 
  Spatial features embedded		& 36.55 \\
  \midrule
  Baseline	(\textit{learnt kernels}) 	& \bfseries 36.62 \\
  Gaussian normalization					& 36.58 \\
  \bottomrule
  \end{tabular*}
  \vspace{-20px}
\end{wraptable}
network to all features, \ie spatial coordinates and grayscale values.
The network learns reasonable features from random initialization, but the PSNR is clearly lower than our baseline despite training $9\times$ longer.

Finally, we evaluate our new normalization approach and learn kernels using scaled basic features. 
Applying a fixed Gaussian filter rather than a flexible, positive kernel for normalization reduces the PSNR by $0.04$dB. 
\subsection{Dense Prediction Tasks}
We next apply our semantic lattice in deep networks for challenging dense prediction tasks, where networks typically operate on downsampled images 
and use bilinear upsampling as a last step, \eg \cite{Chen:2018:EAS,Sun:2018:PCO}.

\begin{figure*}[tb]
  \centering
  \begin{subfigure}[b]{0.328\linewidth}
              \centering
              \includegraphics[width=\linewidth]{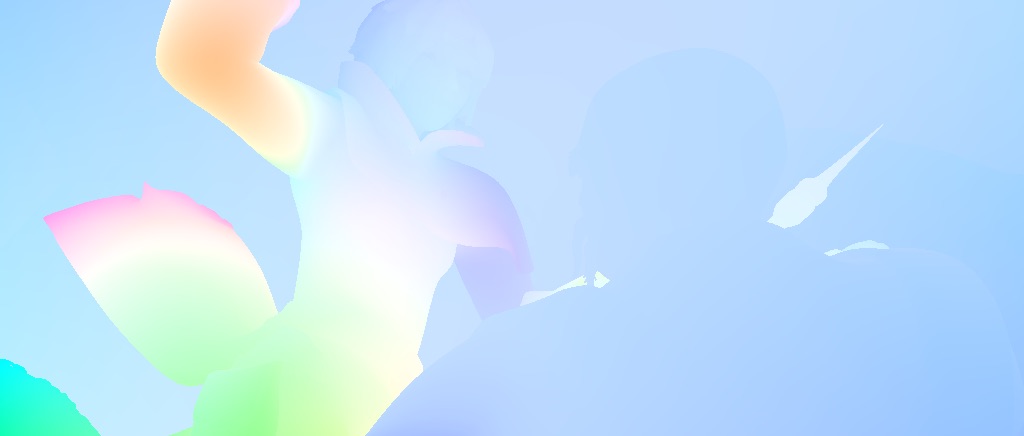}
              \caption{Ground truth flow}
  \end{subfigure}
    \hfill
  \begin{subfigure}[b]{0.328\linewidth}
    \centering
    \includegraphics[width=\linewidth]{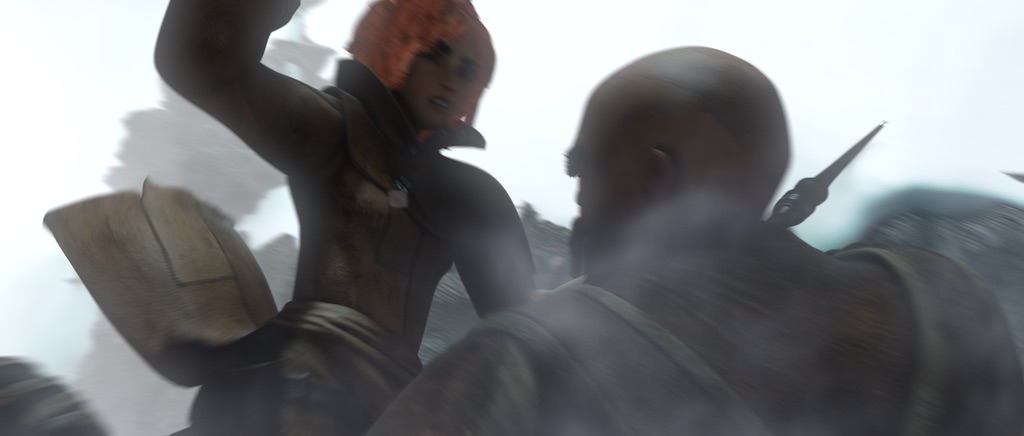}
    \caption{RGB guidance image}
  \end{subfigure}
   \hfill
  \begin{subfigure}[b]{0.328\linewidth}
    \centering
    \includegraphics[width=\linewidth]{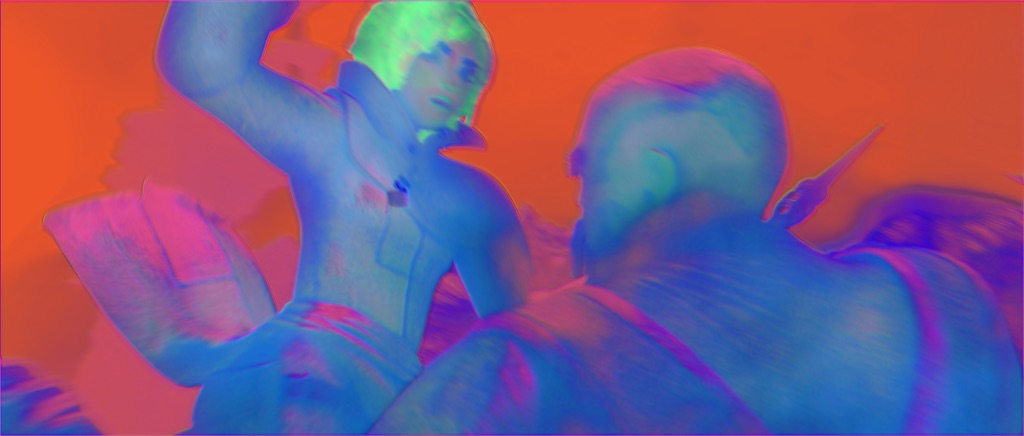}
    \caption{Learnt feature embedding}
  \end{subfigure} \\
  \vspace{0.04cm}
  \begin{subfigure}[b]{0.328\linewidth}
    \centering
    \includegraphics[width=\linewidth]{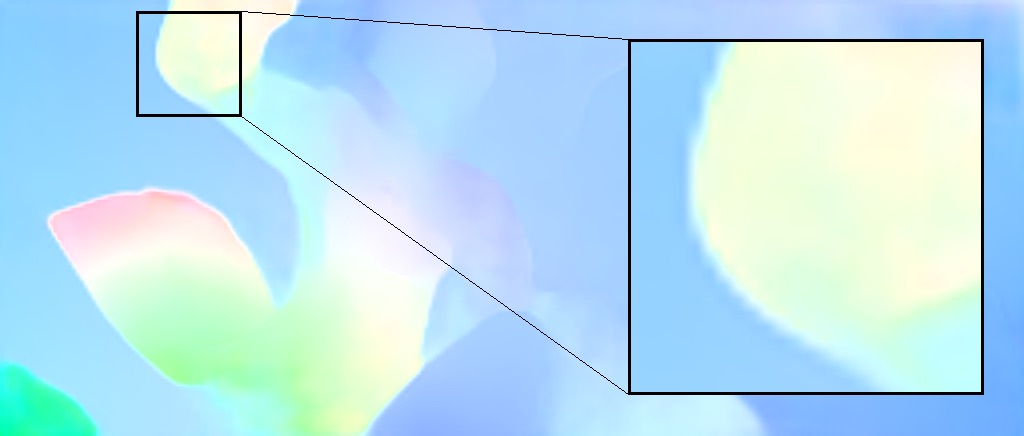}
    \caption{Output PWC-Net \cite{Sun:2018:PCO}}
  \end{subfigure}
  \hfill
   \begin{subfigure}[b]{0.328\linewidth}
              \centering
              \includegraphics[width=\linewidth]{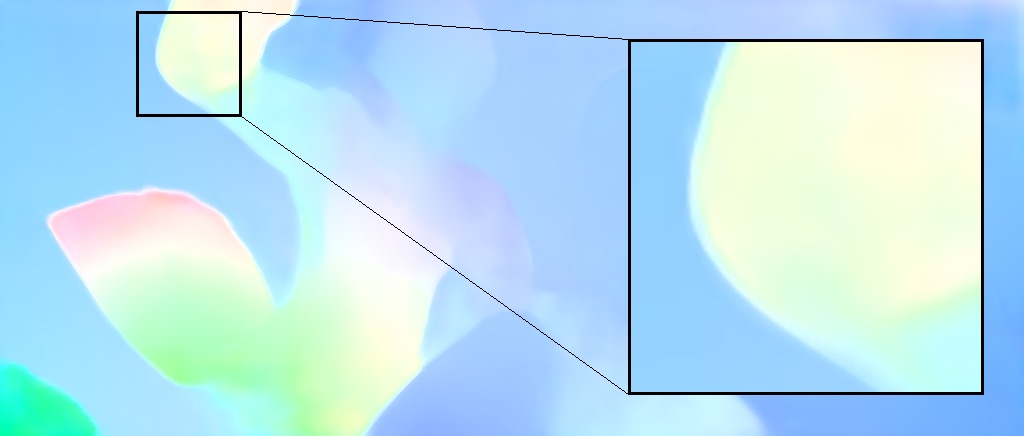}
              \caption{Output DGF \cite{Wu:2018:FET}}
  \end{subfigure}
  \hfill
  \begin{subfigure}[b]{0.328\linewidth}
    \centering
    \includegraphics[width=\linewidth]{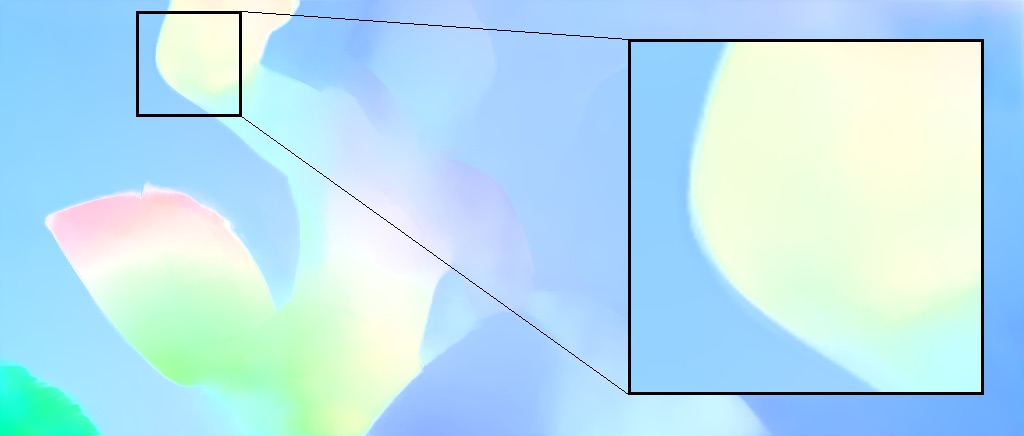}
    \caption{Output semantic lattice}
  \end{subfigure} \\
  \caption{Flow fields and corresponding guidance data for a Sintel sequence.}
  \label{fig:visualization_flow_maps}
  \vspace{-0.5em}
\end{figure*}

\myparagraph{Optical Flow.}
We first consider optical flow for which PWC-Net \cite{Sun:2018:PCO} performs competitively on different benchmarks, \eg \cite{Butler:2012:NOS}.
However, the calculated flow looks blurry and boundary details are oversmoothed, see \cref{fig:visualization_flow_maps}.
We attribute this to the non-adaptive upsampling that enlarges the estimated flow by $\sim 4\times$.

To obtain sharper and more detailed flow, we replace the bilinear upsampling with a single convolution in the semantic lattice.
As basis, we use the so-called \emph{{PWC-Net\_ROB}} model trained on a variety of datasets.
For fair comparison, we do not backpropagate into the network itself but only update the parameters of the semantic lattice, since bilinear upsampling cannot benefit from learning.
Spatial coordinates and color values of the first image are leveraged as basic features. 
The high-resolution guidance image is equally used for input and output features.
Our setup is trained on the Sintel dataset \cite{Butler:2012:NOS}, 
which we split randomly into 862 training, 80 validation, and 99 test images.
We use the average end-point error (AEE) as loss function and train all configurations for 100 epochs on random $281 \times 512$ crops.
Learning rates are set to $1e-3$ and $1e-7$ for embedding parameters and permutohedral kernels, respectively.
We again compare our approach to DGF \cite{Wu:2018:FET}, which we trained for $500$ epochs using their setup for computer vision tasks and hyperparameters tuned on validation.  

\begin{table}[tb]
\centering
  \centering
  \small
  \caption{Average-end-point error (AEE) and boundary AEE (bAEE) on our Sintel test split (ours) and the official Sintel test set (off.). Semantic lattice abbreviated to SL.}
  \label{tab:results_flow}
  \begin{tabularx}{\linewidth}{@{}Xc@{\hskip 1em}c@{\hskip 1em}c@{\hskip 1em}c@{\hskip 1em}cc@{}}
  \toprule
  & \multicolumn{2}{c}{clean (ours)} & \multicolumn{2}{c}{final (ours)} & clean (off.) & final (off.) \\
  \cmidrule(r){2-3}\cmidrule(r){4-5} \cmidrule(lr){6-6}  \cmidrule(lr){7-7}
  & AEE & bAEE  & AEE & bAEE & AEE & AEE\\
  \midrule
  SL (\textit{scaled basic features})	& 1.27 & 7.84 & 1.66	& 8.65 & -- & --	\\
  SL (\textit{learnt embedding})			& 1.26	& 7.50	& 1.66	& 8.61	& -- & --	\\
  SL (\textit{both learnt})					& \bfseries 1.25	& \bfseries 7.49	& \bfseries 1.65 & \bfseries 8.56 & \bfseries 3.84 & \bfseries 4.89 \\
  \midrule
  PWC-Net \cite{Sun:2018:PCO}					& 1.30 & 8.52	& 1.67	& 8.98 & 3.90 & 4.90 \\
  DGF \cite{Wu:2018:FET}							& 1.29 & 8.31	& 1.67 & 8.91 & -- & --\\
  \bottomrule
  \end{tabularx}
\end{table}
\cref{tab:results_flow} shows results on our own test split of Sintel clean and final as well as on the official test images of the benchmark.
Our proposed semantic lattice layer leads to a moderate AEE improvement on both sets.
This is to be expected as our experimental setup can only refine the flow estimates.
However, sharper flow boundaries are clearly visible when considering the results in \cref{fig:visualization_flow_maps}.
As the AEE is known to be insensitive towards boundary accuracy, we also evaluate a boundary average end-point error (bAEE).
It focuses on accuracy close to motion discontinuities, which are determined from ground truth flow by applying a threshold to the flow gradient norm, \cf \cite{Weinzaepfel:2015:LDM}.
As the varying motion ranges require different thresholds \cite{Butler:2012:NOS,Weinzaepfel:2015:LDM},
we follow Weinzaepfel \etal \cite{Weinzaepfel:2015:LDM} and generate multiple masks using values in $\left\{ 1, 3, 7, 10 \right\}$.
These masks are subsequently dilated with a structuring element of size $3$.
We finally calculate the bAEE by evaluating flow on boundary regions only and averaging over the different boundary masks.
Our proposed approach shows a clear benefit for boundary regions, improving the bAEE much more significantly than DGF \cite{Wu:2018:FET} on Sintel clean and final.

\myparagraph{Semantic Segmentation.}
We finally consider the task of semantic segmentation and replace the bilinear upsampling of the recent DeepLabv3+ \cite{Chen:2018:EAS} with our semantic lattice.
Again, we only update parameters of the semantic lattice and keep the remaining network fixed to an XCeption65 model trained on COCO and Pascal VOC augmented, \cf \cite{Chen:2018:EAS}.
Basic features and the setup of our lattice layer remain the same as for optical flow. 
We train the semantic lattice with random crops of size $200 \times 272$ on the training set of Pascal VOC 2012 \cite{Everingham:2015:PVO}, which we further split into training and validation. 
The embedding network is trained for $25$ epochs with a learning rate of $1e-3$, which we reduce by $10\times$ for the remaining $75$ epochs.
The learning rate for permutohedral kernels is fixed to $1e-8$.
DGF is trained as for optical flow with hyperparameters used in \cite{Wu:2018:FET}.

While the semantic lattice without learnt embedding performs slightly worse than the original implementation, the full semantic lattice and DGF outperform
\begin{wraptable}{r}{0.58\textwidth} 
  \centering
  \small
  \caption{Mean intersection over union (mIoU) for \emph{semantic segmentation} on our Pascal VOC 2012 test set.}
  \label{tab:results_segmentation}
  \begin{tabularx}{\linewidth}{@{}Xc@{}}
  \toprule
    & mIoU \\
  \midrule
  Semantic lattice (\textit{scaled basic features})		& 82.17\%	\\
  Semantic lattice (\textit{learnt embedding})			& 82.24\% \\ 
  Semantic lattice (\textit{both learnt})						& 82.25\% \\
  \midrule
  DeepLabv3+ \cite{Chen:2018:EAS} 			& 82.20\% 	\\
  DGF \cite{Wu:2018:FET}							& \bfseries 82.26\%	\\
  \bottomrule
 \end{tabularx}
 \vspace{-20px}
\end{wraptable}
DeepLabv3+.
\cref{tab:results_segmentation} summarizes results on Pascal VOC 2012 validation; see supplemental for visualizations.
The overall improvement is rather small, which we attribute mainly to DeepLabv3+ being highly engineered, with particular focus on the decoder (unlike the previous DeepLabv3).
Nevertheless, image-adaptive filters may benefit further from jointly training with the entire network.
 
\section{Conclusion}
We introduced the semantic lattice layer, a task-specific, generalized convolution.
Our approach is built on the permutohedral lattice that rearranges input data according to different features and thus performs non-local operations with small filter kernels.
First, we generalized the operations in permutohedral space to feature representations that can be learnt from data.
We then showed how rich feature embeddings can be learnt in practice and validated the proposed architecture.
When applying the semantic lattice to color upsampling, learning task-specific features showed a clear benefit.
Adding the semantic lattice to decoders in deep neural networks for optical flow and semantic segmentation allowed to reduce boundary artifacts and improved the accuracy for both tasks.
 
{\small
\bibliographystyle{splncs04}
\bibliography{short,lit}}

\title{Learning Task-Specific Generalized Convolutions in the Permutohedral Lattice \\ {\large -- Supplemental Material --}}
\titlerunning{Learning Task-Specific Generalized Convolutions in the Perm. Lattice}
\authorrunning{Anne S. Wannenwetsch, Martin Kiefel, Peter V. Gehler, Stefan Roth}
\author{
Anne S. Wannenwetsch\inst{1}\thanks{This project was mainly done during an internship at Amazon, Germany.}\orcidID{0000-0002-7016-3820}, \\
Martin Kiefel\inst{2}\orcidID{0000-0001-9432-5428}, \\
Peter V. Gehler\inst{2}\orcidID{0000-0002-5812-4052},
Stefan Roth\inst{1}\orcidID{0000-0001-9002-9832}
}
\institute{$^1$ TU Darmstadt, Germany \qquad $^2$ Amazon, Germany}
\maketitle
\renewcommand{\thesubsection}{\Alph{subsection}}
\setcounter{equation}{6}
\setcounter{figure}{3}
\setcounter{table}{4}

\subsection{Implementation Details}
We implement forward and backward passes of our semantic lattice layer in C++ and CUDA and wrap them in MXNet \citesupp{Chen:2015:MFE}.
To derive the necessary gradients, we manually apply the principles of reverse automatic differentiation, \cf \citesupp{Baydin:2018:ADM}.
Code is available at \url{https://github.com/visinf/semantic_lattice}.

The grid search for scale parameters is performed on full-size training images using the respective evaluation metrics.
If the basic features include color channels, we only determine a single scale factor across all color channels to keep the grid search feasible.
\cref{tab:scale_factors} summarizes scale factors $\lambda_S$ and $\lambda_I$ used in our experiments for spatial and intensity features, respectively.
Remaining hyperparameters, \eg the learning rates, are chosen using the validation set.

We use a default batchsize of $16$ and average over multiple batches if memory permits only fewer samples.
In rare cases, we observe large gradients in training
\begin{wraptable}{r}{0.53\textwidth}
\centering
\small
\caption{Feature scale factors.}
\label{tab:scale_factors}
\begin{tabularx}{\linewidth}{@{}Xc@{\hskip 1.5em}c@{\hskip 0.5em}}
\toprule
Task											& $\lambda_S$ 	å& $\lambda_I$	\\
\midrule
Color upsampling, 2$\times$ 	& 1.25	& 5.0 \\
Color upsampling, 4$\times$ 	& 0.65	& 5.0 \\
Color upsampling, 8$\times$	& 0.20	& 7.5 \\
Optical flow	upsampling			& 0.15	& 70.0 \\
Semantic segmentation ups. 	& 0.15 & 25.0 \\
\bottomrule
\end{tabularx}
\vspace{-20px}
\end{wraptable}
the embedding network, which we counter with gradient clipping (at $0.1$).
The feature network is randomly initialized with a default Xavier initialization \citesupp{Glorot:2010:UDT} while the filter weights in permutohedral space are initialized as Gaussian kernels.
The non-negativity of normalization filters is ensured by learning in the log-domain.
 
\subsection{Network architectures}
The architecture of our embedding networks is described in \cref{tab:embedding_architecture}.
The parameter $\tilde{d}$ denotes the number of embedded features.
We set $\tilde{d}=1$ for color upsampling and $\tilde{d}=3$ for dense prediction tasks.
Since embedded features are concatenated with two-dimensional spatial coordinates, 
the semantic lattice receives features of dimensionality $d=\tilde{d}+2$.
All convolution layers use a stride of one and zero padding to preserve the input size.
We apply leakyReLU activations with slope coefficients $\alpha = 0.2$ as non-linearities.
For batch normalization, we use default parameters and apply the transformation to the channel axis.

\cref{tab:lattice_architecture} specifies the setup in permutohedral space used for our experiments.
The number of outputs $c$ depends on the specific task and is determined by the dimensionality of the input data. 
As such, we have $c=3$ for color upsampling, $c=2$ for optical flow and $c=21$ for semantic segmentation.
All convolutions in the permutohedral lattice are performed per channel, \ie we set the number of groups to $c$ and learn a separate convolution kernel for each data dimension.

\begin{table}[tb]
\begin{minipage}[c]{0.55\linewidth}
\centering
\small
\caption{Architecture of embedding networks.}
\label{tab:embedding_architecture}
\begin{tabularx}{\linewidth}{@{}Xc@{\hskip 2em}c@{\hskip 2em}c@{\hskip 1.5em}}
\toprule
Layer						& 1 			& 2 			& 3	\\
\midrule
Kernel size				& 3			& 3 			& 3	\\ 
Channels 					& 15			& 15			& $\tilde{d}$ \\ 
Groups						& 1			& 1			& 1 \\
Bias							& \cmark	& \cmark & \cmark \\
Non-linearity 			& \cmark	& \cmark	& \xmark \\
Batch normalization	& \xmark	& \xmark	& \cmark \\
\bottomrule
\end{tabularx}
\end{minipage}
\hfill
\begin{minipage}[c]{0.42\linewidth}
\centering
\small
\caption{Architecture in perm. space.}
\label{tab:lattice_architecture}
\begin{tabularx}{\linewidth}{@{}Xc@{\hskip 1.5em}}
\toprule
Layer						& 1 \\
\midrule
Neighborhood size	& 3	\\ 
Channels 					& $c$ \\
Groups						& $c$ \\
Bias							& \xmark \\
Non-linearity 			& \xmark \\
Batch normalization	& \xmark \\
\bottomrule
\end{tabularx}
\end{minipage}
\end{table} 
\subsection{Additional Experiments Color Upsampling}
We start with a small ablation study performed on the task of color upsampling.
In a first experiment, we keep the permutohedral weights fixed and increase the number of embedded features from one to two.
However, the additional feature does not lead to improved results but we even observe a small drop in PSNR on the test split from 36.81 to 36.79.
As a larger amount of feature dimensions leads to an increased runtime, we choose the dimensionality of the feature embedding to equal the number of basic features provided to the embedding network.

In a second setup, we evaluate the performance of the semantic lattice with larger kernels in permutohedral space.
Therefore, we learn permutohedral weights of neighborhood size two with fixed basic features.
We obtain a PSNR of 36.64 on the test set in comparison to a PSNR of 36.62 with a neighborhood size of one.
Again, we choose the smaller neighborhood size due to improved runtime.

Finally, we evaluate color upsampling for additional upsampling factors in \cref{tab:additional_results_colorization}. We again observe that the semantic lattice performs best if the feature embedding as well as the kernels are learnt. Interestingly, the benefit gets more significant as the difficulty of the task increases, \ie for larger upsampling factors. As before, we clearly outperform baselines and related work.
\begin{table}[tb]
  \centering
  \small
  \caption{Evaluation of additional upsampling factors for the task of \emph{color upsampling} on the Pascal VOC 2012 Segmentation test set. $^*$No pretrained network available.}
  \label{tab:additional_results_colorization}
  \begin{tabularx}{\textwidth}{@{}Xc@{\hskip 1.5em}c@{}}
  \toprule
   & PSNR  [dB], $\times 2$ & PSNR  [dB], $\times 8$ \\
  \midrule
  Semantic lattice (\textit{scaled basic features})	& 40.05 & 33.93 \\
  Semantic lattice (\textit{learnt kernels})				& 40.10 & 34.06 \\ 
  Semantic lattice (\textit{learnt embedding})		& 40.20 & 34.23	\\  
  Semantic lattice (\textit{both learnt})					& \bfseries 40.22 & \bfseries 34.33 \\
  \midrule
  Nearest neighbors				& 25.91  & 19.46 \\
  Bicubic upsampling				& 27.23  & 20.73 \\
  DGF \cite{Wu:2018:FET}	& 37.80  & 32.97 \\
  DJIF \cite{Li:2019:JIF}		& --$^*$ & 20.57 \\
  \bottomrule 
 \end{tabularx}
\end{table} 
\subsection{Additional Visualizations}

We provide visualizations for the different tasks discussed in the paper.

\begin{figure*}[tb]
  \centering
  \begin{subfigure}[b]{0.328\linewidth}
              \centering
              \includegraphics[width=\linewidth]{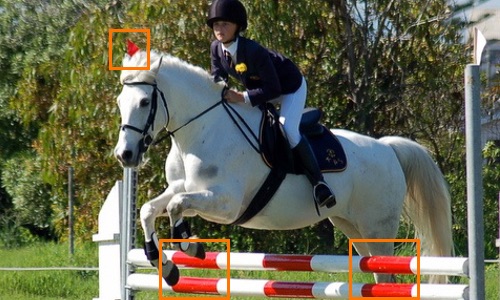}
  \end{subfigure}
    \hfill
  \begin{subfigure}[b]{0.328\linewidth}
    \centering
    \includegraphics[width=\linewidth]{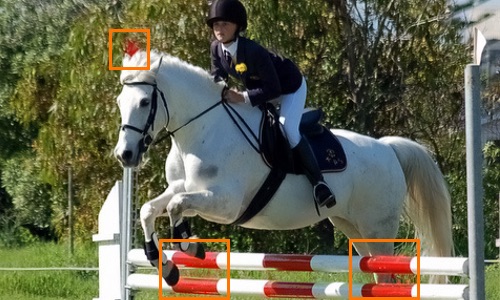}
  \end{subfigure}
   \hfill
  \begin{subfigure}[b]{0.328\linewidth}
    \centering
    \includegraphics[width=\linewidth]{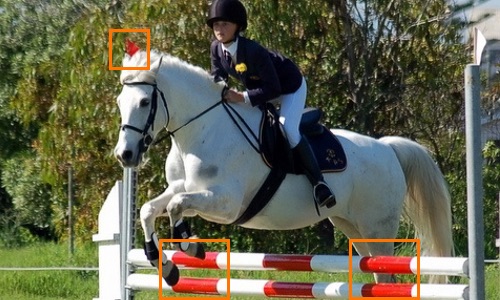}
  \end{subfigure} \\
  \vspace{0.08cm}
  \begin{subfigure}[b]{0.328\linewidth}
    \centering
    \includegraphics[width=\linewidth]{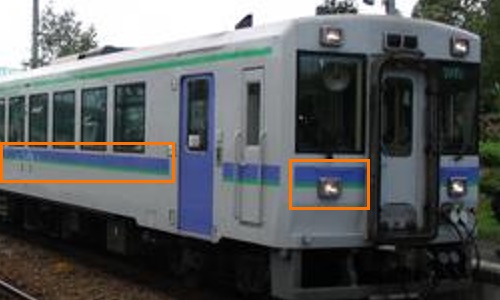}
  \end{subfigure}
  \hfill
   \begin{subfigure}[b]{0.328\linewidth}
              \centering
              \includegraphics[width=\linewidth]{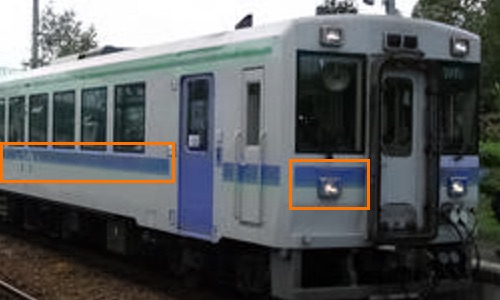}
  \end{subfigure}
  \hfill
  \begin{subfigure}[b]{0.328\linewidth}
    \centering
    \includegraphics[width=\linewidth]{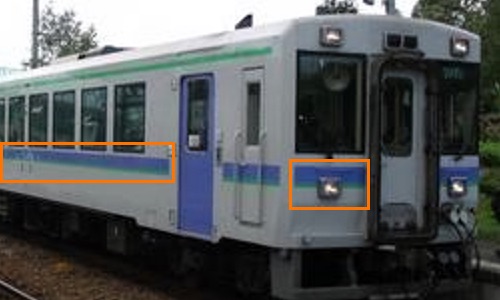}
  \end{subfigure} \\
    \vspace{0.08cm}
  \begin{subfigure}[b]{0.328\linewidth}
    \centering
    \includegraphics[width=\linewidth]{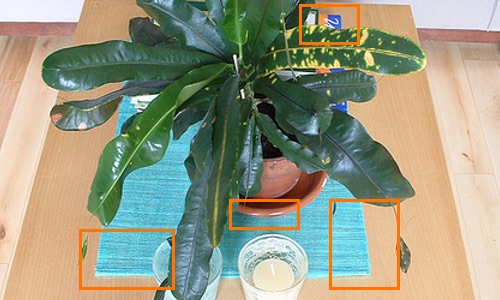}
  \end{subfigure}
  \hfill
   \begin{subfigure}[b]{0.328\linewidth}
              \centering
              \includegraphics[width=\linewidth]{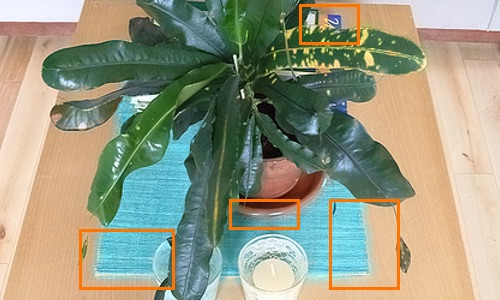}
  \end{subfigure}
  \hfill
  \begin{subfigure}[b]{0.328\linewidth}
    \centering
    \includegraphics[width=\linewidth]{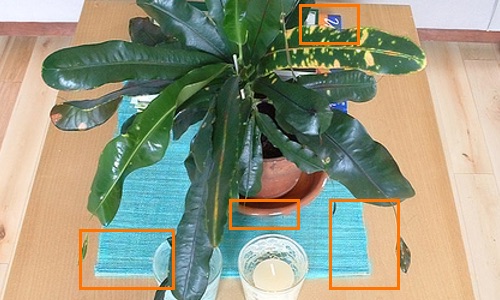}
  \end{subfigure} \\
      \vspace{0.08cm}
  \begin{subfigure}[b]{0.328\linewidth}
    \centering
    \includegraphics[width=\linewidth]{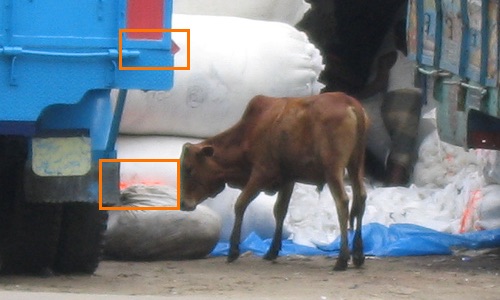}
  \end{subfigure}
  \hfill
   \begin{subfigure}[b]{0.328\linewidth}
              \centering
              \includegraphics[width=\linewidth]{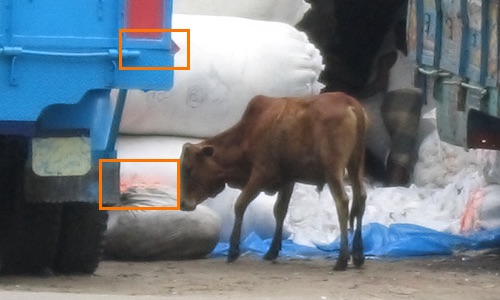}
  \end{subfigure}
  \hfill
  \begin{subfigure}[b]{0.328\linewidth}
    \centering
    \includegraphics[width=\linewidth]{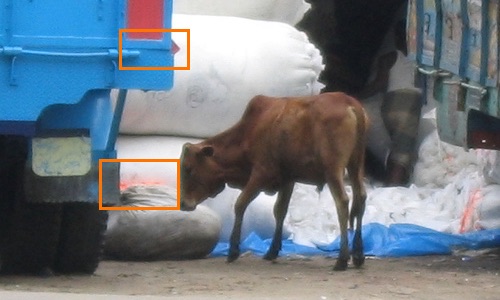}
  \end{subfigure} \\
  \caption{Left to right: Crops of ground truth, outputs of DGF \cite{Wu:2018:FET}, and outputs of the semantic lattice for $4 \times$ color upsampling on Pascal VOC 2012. Best viewed on screen.}
  \label{fig:additional_visualizations_color}
  \vspace{-0.5em}
\end{figure*}

\myparagraph{Color upsampling.} 
Visualizations of the $4 \times$ color upsampling task are given in \cref{fig:additional_visualizations_color}.
The fully learnt semantic lattice is clearly superior to the deep guided filter as it correctly reconstructs small and thin color regions, \eg the green and blue strips on the white train. Moreover, the lattice shows considerably fewer color bleeding artifacts, \eg at the red parts of the bars in the first row.

\begin{figure*}[tb]
  \centering
  \begin{subfigure}[b]{0.328\linewidth}
              \centering
              \includegraphics[width=\linewidth]{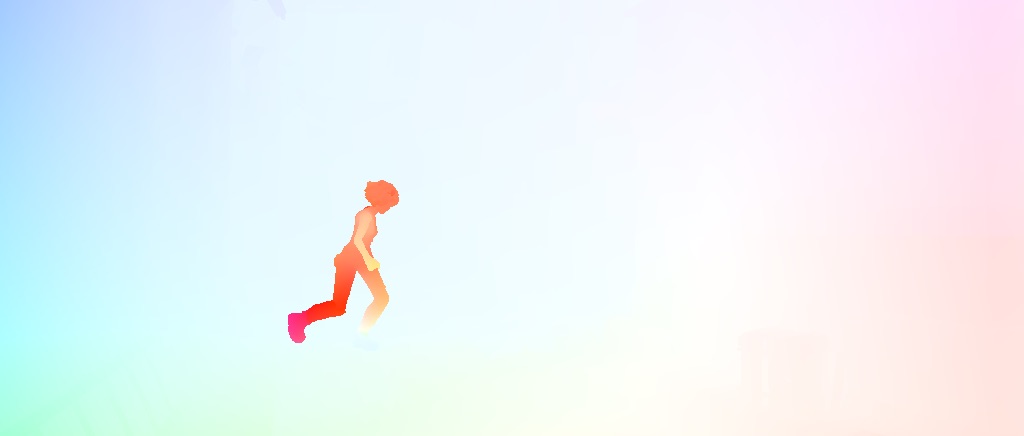}
  \end{subfigure}
    \hfill
  \begin{subfigure}[b]{0.328\linewidth}
    \centering
    \includegraphics[width=\linewidth]{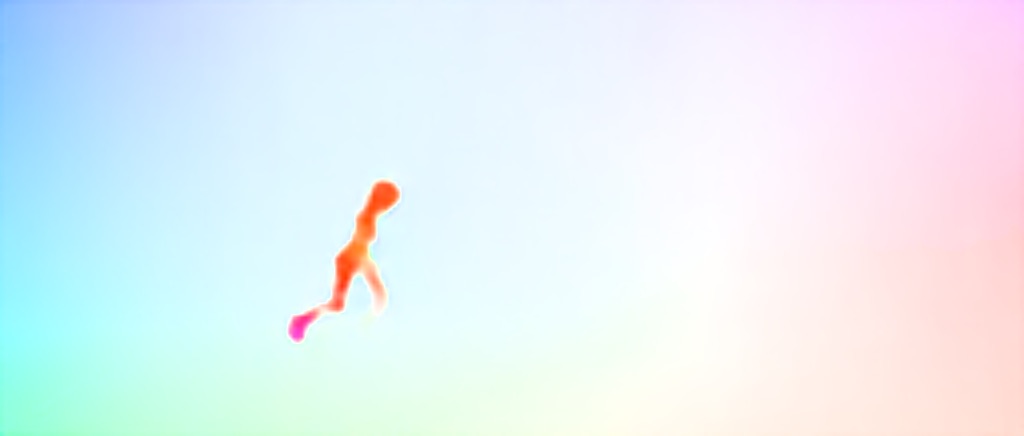}
  \end{subfigure}
   \hfill
  \begin{subfigure}[b]{0.328\linewidth}
    \centering
    \includegraphics[width=\linewidth]{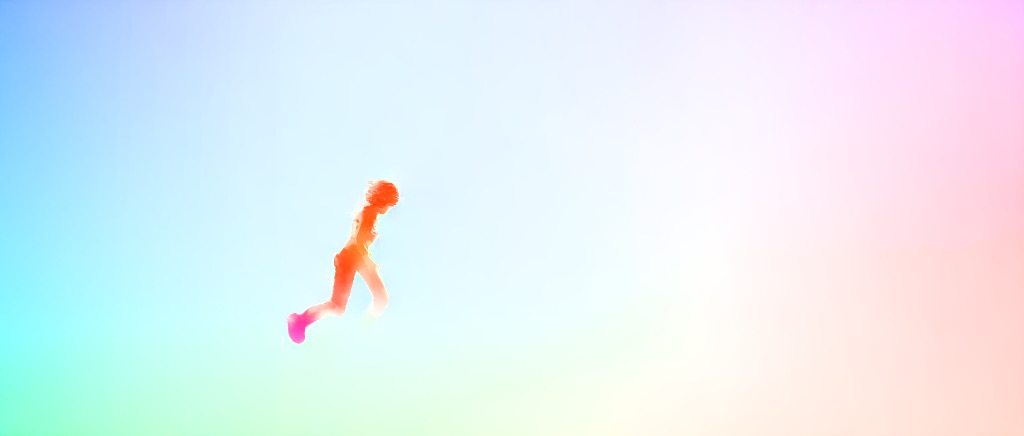}
  \end{subfigure} \\
  \vspace{0.08cm}
  \begin{subfigure}[b]{0.328\linewidth}
    \centering
    \includegraphics[width=\linewidth]{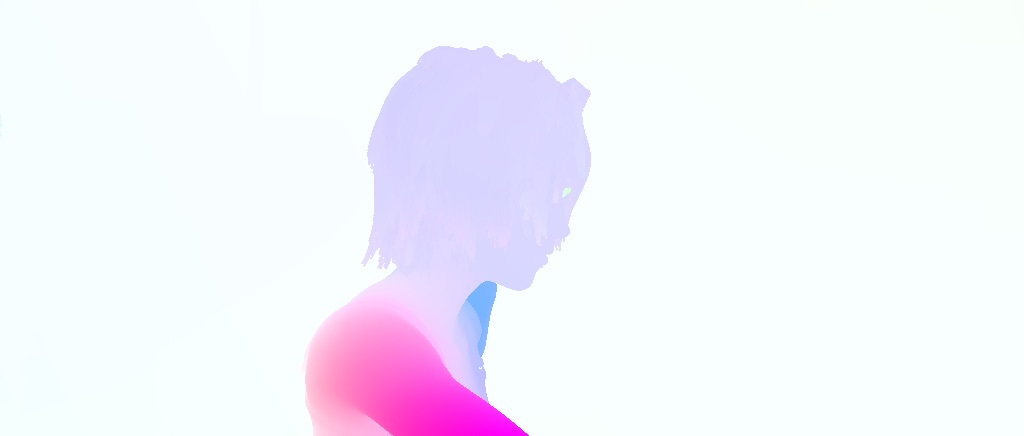}
  \end{subfigure}
  \hfill
   \begin{subfigure}[b]{0.328\linewidth}
              \centering
              \includegraphics[width=\linewidth]{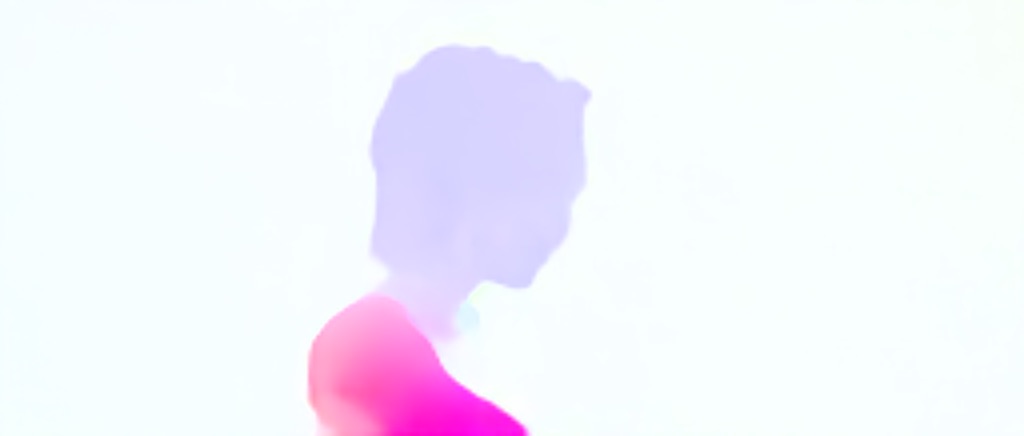}
  \end{subfigure}
  \hfill
  \begin{subfigure}[b]{0.328\linewidth}
    \centering
    \includegraphics[width=\linewidth]{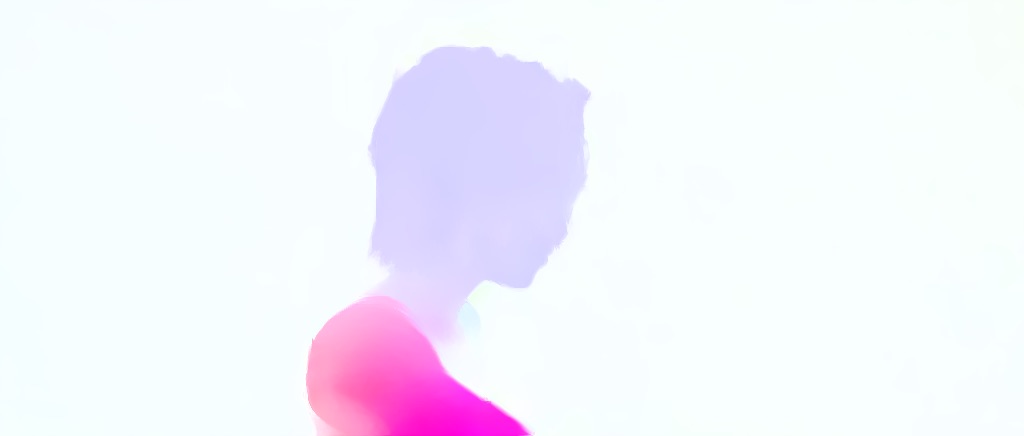}
  \end{subfigure} \\
    \vspace{0.08cm}
  \begin{subfigure}[b]{0.328\linewidth}
    \centering
    \includegraphics[width=\linewidth]{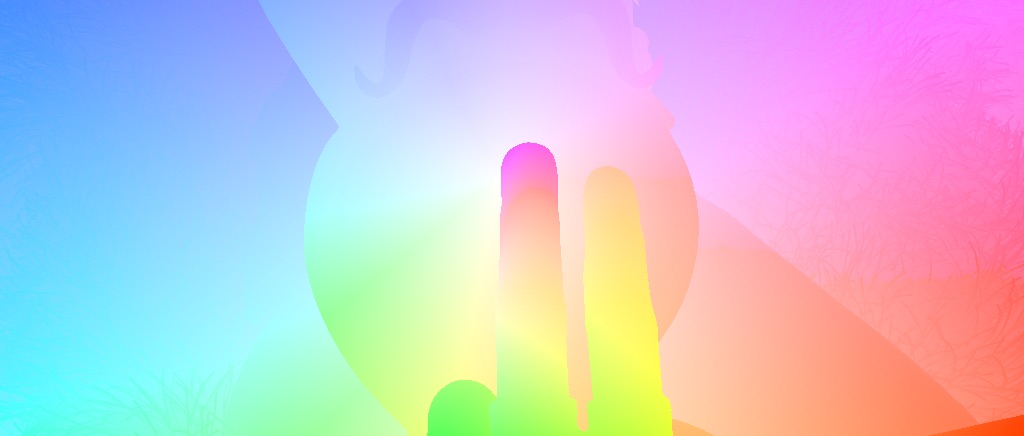}
  \end{subfigure}
  \hfill
   \begin{subfigure}[b]{0.328\linewidth}
              \centering
              \includegraphics[width=\linewidth]{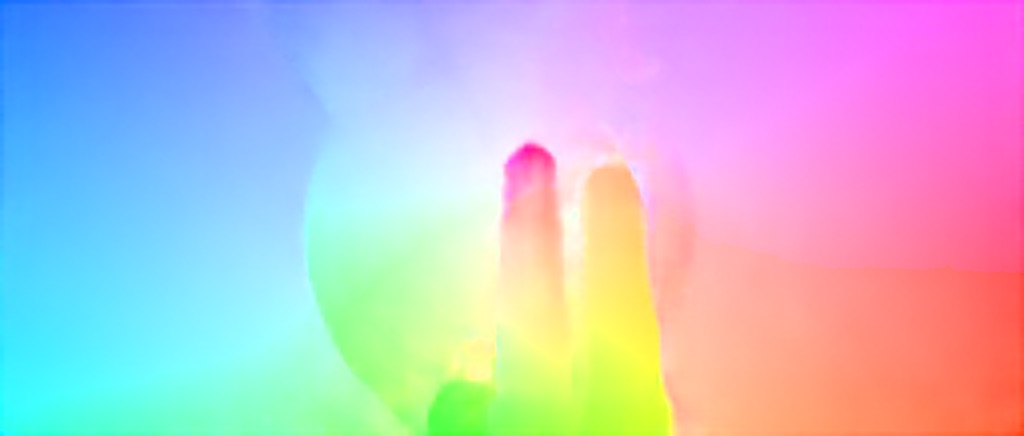}
  \end{subfigure}
  \hfill
  \begin{subfigure}[b]{0.328\linewidth}
    \centering
    \includegraphics[width=\linewidth]{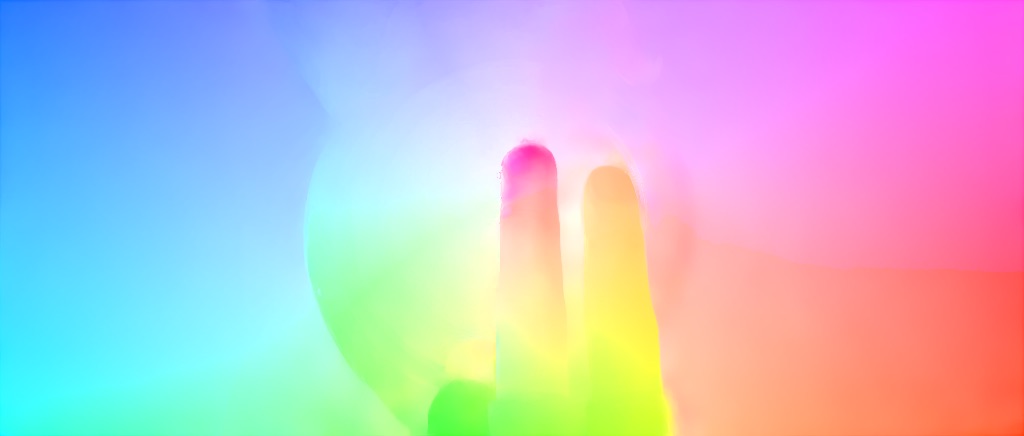}
  \end{subfigure} \\
      \vspace{0.08cm}
  \begin{subfigure}[b]{0.328\linewidth}
    \centering
    \includegraphics[width=\linewidth]{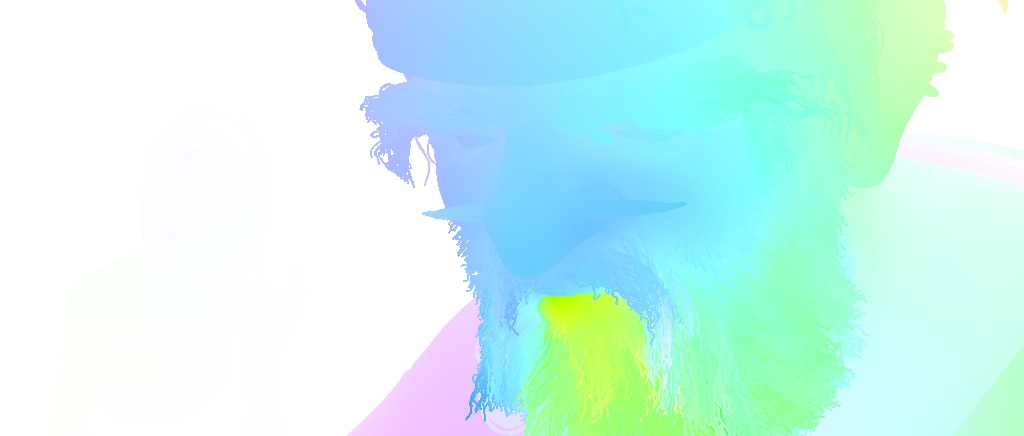}
  \end{subfigure}
  \hfill
   \begin{subfigure}[b]{0.328\linewidth}
              \centering
              \includegraphics[width=\linewidth]{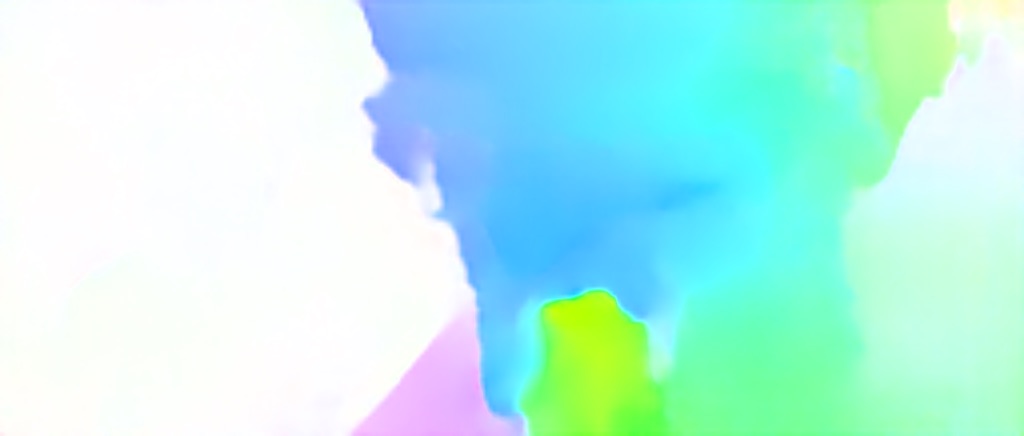}
  \end{subfigure}
  \hfill
  \begin{subfigure}[b]{0.328\linewidth}
    \centering
    \includegraphics[width=\linewidth]{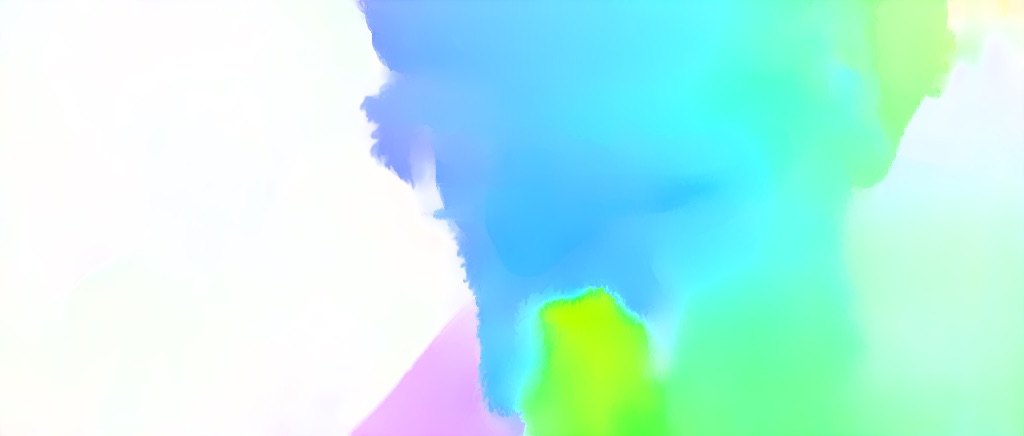}
  \end{subfigure} \\
  \caption{Left to right: Ground truth, outputs of PWC-Net \cite{Sun:2018:PCO}, and outputs of the fully learnt semantic lattice on different Sintel sequences. Best viewed on screen.}
  \label{fig:additional_visualizations_flow}
  \vspace{-0.5em}
\end{figure*}

\myparagraph{Dense prediction tasks.}
\cref{fig:additional_visualizations_flow} shows visualizations for ground truth and predicted optical flow on several sequences of the Sintel dataset \cite{Butler:2012:NOS}.
As already discussed in the main paper, the semantic lattice leads to less blurry flow fields in comparison to the original PWC-Net \cite{Sun:2018:PCO}. Additionally, it allows to recover fine details at motion boundaries, \eg the structure of the hair in the last row.

\begin{figure*}[tb]
  \centering
  \begin{subfigure}[b]{0.328\linewidth}
              \centering
              \includegraphics[width=\linewidth]{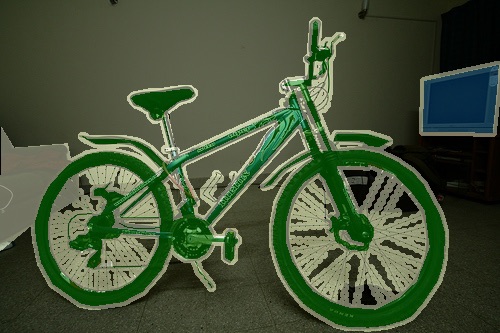}
  \end{subfigure}
    \hfill
  \begin{subfigure}[b]{0.328\linewidth}
    \centering
    \includegraphics[width=\linewidth]{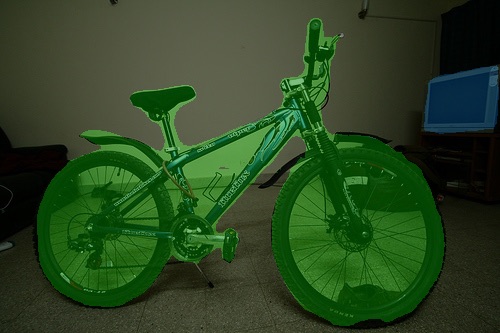}
  \end{subfigure}
   \hfill
  \begin{subfigure}[b]{0.328\linewidth}
    \centering
    \includegraphics[width=\linewidth]{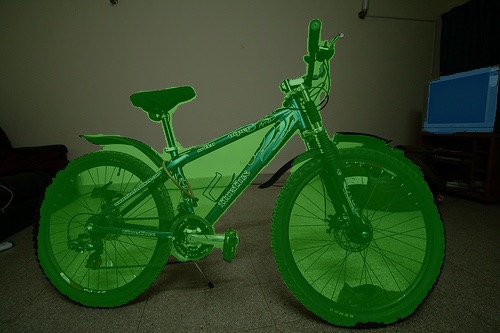}
  \end{subfigure} \\
  \vspace{0.08cm}
  \begin{subfigure}[b]{0.328\linewidth}
    \centering
    \includegraphics[width=\linewidth]{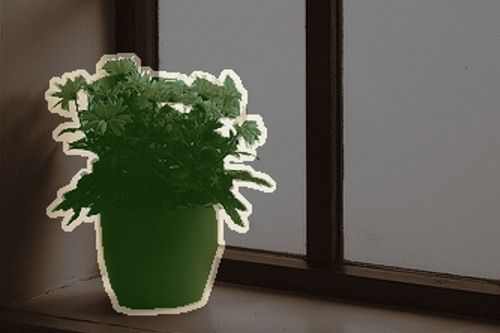}
  \end{subfigure}
  \hfill
   \begin{subfigure}[b]{0.328\linewidth}
              \centering
              \includegraphics[width=\linewidth]{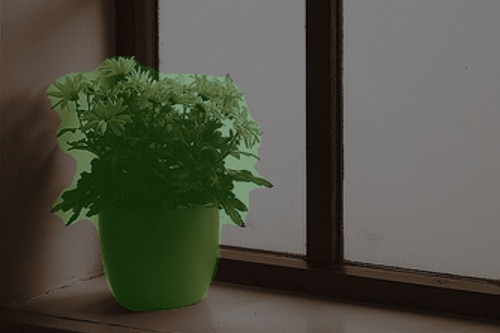}
  \end{subfigure}
  \hfill
  \begin{subfigure}[b]{0.328\linewidth}
    \centering
    \includegraphics[width=\linewidth]{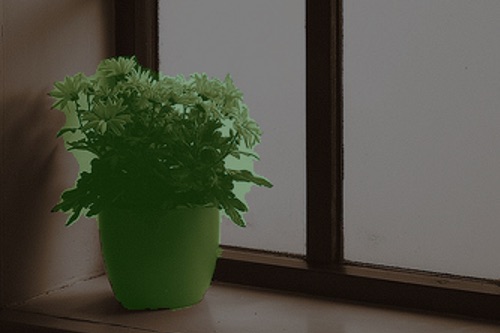}
  \end{subfigure} \\
    \vspace{0.08cm}
  \begin{subfigure}[b]{0.328\linewidth}
    \centering
    \includegraphics[width=\linewidth]{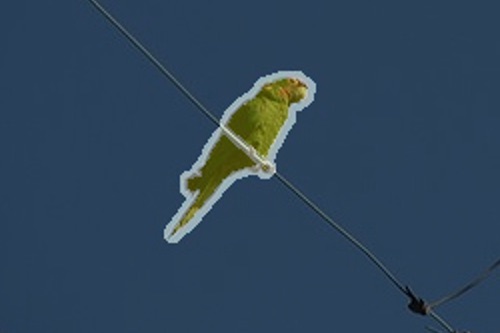}
  \end{subfigure}
  \hfill
   \begin{subfigure}[b]{0.328\linewidth}
              \centering
              \includegraphics[width=\linewidth]{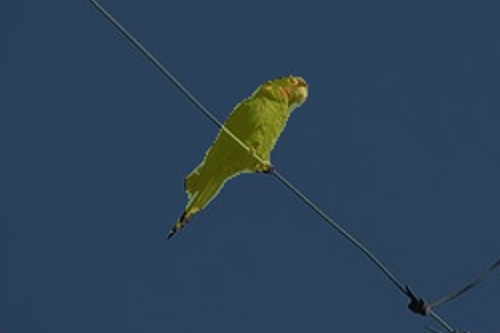}
  \end{subfigure}
  \hfill
  \begin{subfigure}[b]{0.328\linewidth}
    \centering
    \includegraphics[width=\linewidth]{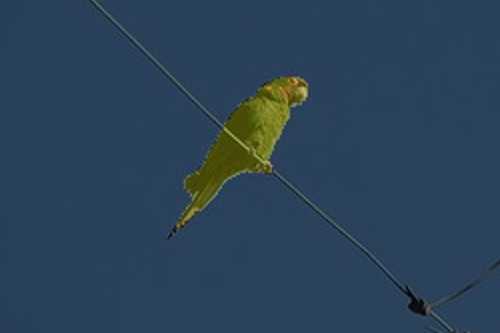}
  \end{subfigure} \\
      \vspace{0.08cm}
  \begin{subfigure}[b]{0.328\linewidth}
    \centering
    \includegraphics[width=\linewidth]{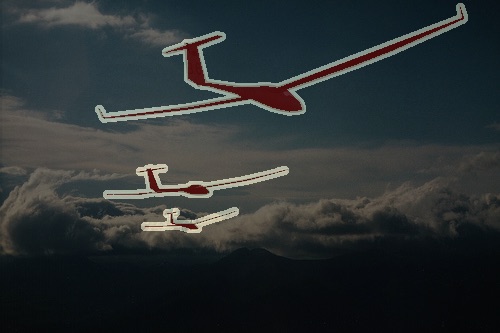}
  \end{subfigure}
  \hfill
   \begin{subfigure}[b]{0.328\linewidth}
              \centering
              \includegraphics[width=\linewidth]{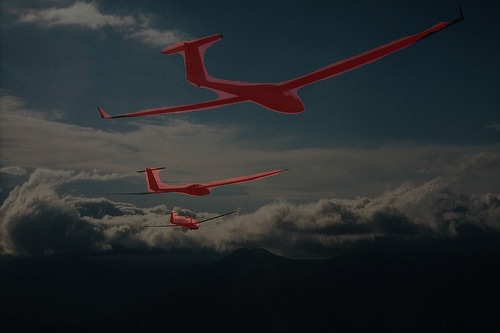}
  \end{subfigure}
  \hfill
  \begin{subfigure}[b]{0.328\linewidth}
    \centering
    \includegraphics[width=\linewidth]{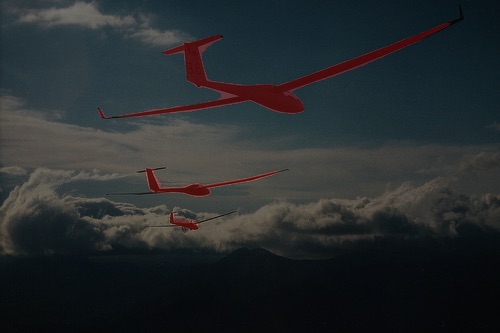}
  \end{subfigure} \\
  \caption{Left to right: Crops of ground truth, outputs of DeepLabv3+ \cite{Chen:2018:EAS}, and outputs of the semantic lattice for segmentation on Pascal VOC 2012. Best viewed on screen.}
  \label{fig:additional_visualizations_segmentation}
  \vspace{-0.5em}
\end{figure*}

In \cref{fig:additional_visualizations_segmentation}, examples for segmentations on Pascal VOC 2012 \cite{Everingham:2015:PVO} are provided. 
Considering the results of DeepLabv3+, one observes that the segmentation masks frequently exceed the borders of detected objects.
Applying the semantic lattice with learnt embedding and kernels allows us to reduce such margins. As such, the obtained segmentations align better with the underlying objects.  
{\small
\bibliographystylesupp{splncs04}
\bibliographysupp{short,lit_supp}

\end{document}